# IDU: Incremental Dynamic Update of Existing 3D Virtual Environments with New Imagery Data

**Meida Chen, Luis Leal, Yue Hu, Rong Liu, Butian Xiong, Andrew Feng**
USC Institute for Creative Technologies
Los Angeles, California
{mechen, lleal, yuhu, roliu, bxiong, feng}@ict.usc.edu

**Jiuyi Xu, Yangming Shi**
Colorado School of Mines
Golden, Colorado
{jiuyi_xu, yangming.shi}@mines.edu

## ABSTRACT

For simulation and training purposes, military organizations have made substantial investments in developing high-resolution 3D virtual environments through extensive imaging and 3D scanning. However, the dynamic nature of battlefield conditions—where objects may appear or vanish over time—makes frequent full-scale updates both time-consuming and costly. In response, we introduce the Incremental Dynamic Update (IDU) pipeline, which efficiently updates existing 3D reconstructions, such as 3D Gaussian Splatting (3DGS), with only a small set of newly acquired images. Our approach starts with camera pose estimation to align new images with the existing 3D model, followed by change detection to pinpoint modifications in the scene. A 3D generative AI model is then used to create high-quality 3D assets of the new elements, which are seamlessly integrated into the existing 3D model. The IDU pipeline incorporates human guidance to ensure high accuracy in object identification and placement, with each update focusing on a single new object at a time. Experimental results confirm that our proposed IDU pipeline significantly reduces update time and labor, offering a cost-effective and targeted solution for maintaining up-to-date 3D models in rapidly evolving military scenarios.

## ABOUT THE AUTHORS

**Meida Chen** is currently a Research Scientist at USC-ICT. He received his Ph.D. degree from the University of Southern California. His research focuses on the applications of 3D computer vision for the creation of virtual environments training and simulation. He received the best paper award at IITSEC simulation subcommittee and published numerous papers in top conferences/journals in both the computer vision and civil engineering fields.

**Luis Leal** is a Data Collection Lead at USC-ICT with over a decade of experience in 3D modeling, simulation, and GIS. He has spent more than eight years supporting Department of Defense operations, combining technical expertise with field experience to connect research innovation with mission success.

**Yue Hu** is currently a Graduate Research Assistant and Ph.D. candidate in Computer Engineering at the University of Southern California. Her research focuses on the intersection of Computer Vision and Machine Learning, with applications in robotic perception and intelligent scene understanding. She specializes in 3D vision and geometry, including 3D Gaussian Splatting, NeRF, SLAM, SfM, and relighting, as well as 2D object detection and synthetic dataset generation using Unreal Engine.

**Rong Liu** is currently a Research Engineer at the USC-ICT. He received his Master degree in Computer Science from the University of Southern California. His research lies in the intersection of Computer Vision, Computer Graphics, and Machine Learning.

**Butian Xiong** is currently an MSCS student at USC and a Research Intern at USC-ICT supervised by Andrew Feng. His research focuses on 3D representation, feature lifting, 3D scene understanding, indoor navigation, and large-scale scene reconstruction.

**Andrew Feng** is currently the Associate Director under the Geospatial Terrain Research group at USC-ICT. Dr. Feng holds a PhD and MS in Computer Science from the University of Illinois at Urbana-Champaign. He joined ICT in 2011 as a Research Programmer specializing in computer graphics, then as a Research Associate focusing on 3D avatar generation and motion synthesis, before becoming a Research Computer Scientist within the Geospatial Terrain Research group.

**Jiuyi Xu** is currently a PhD student in Robotics at the Colorado School of Mines, supervised by Dr. Yangming Shi. His research interests lie in Generative AI and Vision-Language-Action (VLA) models.

**Yangming Shi** is currently an assistant professor in Department of Civil and Environmental Engineering and a faculty member in the robotics program at the Colorado School of Mines. He received his Ph.D. from the University of Florida. His research focuses on Human-computer Interaction, Human-robot collaboration and NeuroErgonomics.

# IDU: Incremental Dynamic Update of Existing 3D Virtual Environments with New Imagery Data

**Meida Chen, Luis Leal, Yue Hu, Rong Liu, Butian Xiong, Andrew Feng**
USC Institute for Creative Technologies
Los Angeles, California
{mechen, lleal, yuhu, roliu, bxiong, feng}@ict.usc.edu

**Jiuyi Xu, Yangming Shi**
Colorado School of Mines
Golden, Colorado
{jiuyi_xu, yangming.shi}@mines.edu

**Introduction**

Accurate, up-to-date 3D virtual environments are essential assets in military simulation and training, providing realistic scenarios that enhance operational preparedness and support effective mission planning. Military organizations have extensively invested in sophisticated methods for generating detailed, semantically enriched virtual terrains, driven by advancements in UAV-based photogrammetric 3D reconstruction, and semantic/instance segmentation techniques. Our previous research, consistently presented at the Interservice/Industry Training, Simulation, and Education Conference (I/ITSEC) from 2019 through 2024, has made substantial progress in addressing key challenges associated with creating simulation-ready 3D terrain datasets on scale (Chen et al., 2019a, 2020a, 2020b, 2021a; McCullough et al., 2020; Xu et al., 2024). By leveraging UAV imagery coupled with semantic segmentation techniques, we successfully established robust pipelines for reconstructing large-scale environments that integrate precise semantic context, elevating the fidelity and utility of these virtual terrains for modeling, simulation, and training(Chen et al., 2019b, 2020c, 2020d, 2021b).

Despite these advancements, real-world military scenarios are inherently dynamic, with elements frequently appearing, moving, or vanishing entirely. These constant changes rapidly render simulated 3D environments outdated, necessitating frequent updates to preserve realism and maintain effectiveness. While some formats, such as the Common Database (CDB), support tile-based updates, they typically operate at a coarser granularity and still involve reprocessing a terrain tile that needs to be updated. In contrast, many real-world changes are highly localized—such as the addition of individual objects—which do not warrant full-tile regeneration. Traditionally, even such small changes have required full or tile-level reconstruction, a process that remains time-consuming and resource-intensive. This highlights a clear need for efficient methodologies capable of updating existing 3D models of terrain at a finer spatial resolution, without repeated exhaustive reprocessing.

In response to this critical need, and within the scope of our research focus, we propose a targeted solution explicitly designed to integrate new, localized changes (e.g., placement of barriers) from new imagery into pre-existing 3D reconstructions, and for brevity, we named it Incremental Dynamic Update (IDU) pipeline. Specifically, the IDU pipeline operates through a structured, step-by-step process: (1) 3D reconstruction of the terrain. (2) aligning new images with existing 3D models through precise camera pose estimation (Incremental); (3) systematically detecting and isolating scene modifications (Dynamic); and (4) creating high-quality 3D representations of newly introduced objects, and (Update) the baseline model from step zero. Importantly, each update is carefully managed on a per-object basis, involving human guidance to guarantee accuracy in object identification, spatial placement, and visual coherence within the existing 3D terrain model.

In this paper, we validate the efficacy of the proposed IDU pipeline through experimental analysis using two specially curated datasets of two sites. Specifically, we first collected thousands of aerial images to create baseline 3D virtual environments for both sites. Following this, new objects—such as tents, road barriers, and sandbags—were physically introduced into the environments to simulate realistic changes. Subsequently, a smaller set of images reflecting these updates were captured. We then evaluated the performance of our IDU pipeline by visually comparing the incrementally updated 3D terrains against new ground-truth imagery. Our results demonstrate that the IDU approach reduces the time, manual effort, and computational resources required to maintain dynamic 3D environments. Building directly upon our established expertise in large-scale terrain reconstruction with 3D Gaussian Splatting, the IDU pipeline offers a practical, cost-effective method for preserving simulation realism, rapidly adapting to evolving conditions, and enhancing readiness for complex military settings. An overview of the full IDU pipeline is shown in the video available at: https://youtu.be/efUs3HU_IaY

**Background - 3D reconstruction**

**Traditional Photogrammetric 3D Reconstruction from Aerial Imagery**

Photogrammetry has become a widely adopted approach for generating accurate, detailed 3D reconstructions of real-world environments, particularly in military training and simulation contexts. At its core, photogrammetry can produce reliable three-dimensional representations from overlapping two-dimensional images, typically collected by Unmanned Aerial Vehicles (UAVs) for large terrains (Spicer et al., 2016). Modern photogrammetric workflows generally include several key steps: (1) aerial image acquisition with sufficient image overlap, (2) identifying corresponding tie points across images, (3) solving for camera parameters, and (4) generating dense 3D point clouds with subsequently-derived products such as textured 3D meshes or Digital Surface Models (DSMs) (Chen et al., 2019a).

Commercial photogrammetry software, such as ContextCapture (iTwin), Pix4D, and Metashape, provide established solutions that automate much of this 3D reconstruction process, delivering high-quality, geospatially accurate 3D meshes ready for integration into simulation environments. Although photogrammetry remains a robust and widely used technique for reconstructing the geometry of real-world environments, it does have certain limitations. Specifically, photogrammetric 3D meshes typically have diminished photorealistic rendering capabilities, have difficulty capturing thin structures (e.g., power lines), and struggle to accurately represent reflective or transparent surfaces such as glass or water.

**3D Gaussian Splatting: A State-of-the-Art 3D Scene Representation**

Recent advancements in neural rendering have introduced innovative methods that address some of the limitations from traditional photogrammetric reconstruction techniques (Kerbl et al., 2023; Barron, Mildenhall, Tancik, et al., 2021; Barron, Mildenhall, Verbin, et al., 2023; Martin-Brualla et al., 2021; Mildenhall et al., 2021; Garbin et al., 2021). Among these methods, 3D Gaussian Splatting (3DGS) has emerged as a state-of-the-art approach, particularly notable for its superior rendering quality compared to conventional mesh-based representations (Kerbl et al., 2023). Unlike traditional approaches, 3DGS explicitly models a 3D scene using numerous oriented, anisotropic Gaussian primitives, known as "splats," each encoding spatial position, color, transparency, and shape information. This explicit representation allows 3DGS to produce photorealistic renderings, significantly enhancing visual realism and thus holding substantial promise for improving simulation realism in military training scenarios (Chen et al., 2024).

Our decision to adopt 3DGS for the IDU pipeline is motivated primarily by its superior rendering capabilities, which significantly enhance visual fidelity in the training simulation environments. This capability results from its differentiable volumetric rendering approach, wherein the attributes of the Gaussian primitives are optimized directly through a neural rendering pipeline. Moreover, the differentiable rendering nature of the 3DGS could facilitate camera pose refinement by optimizing camera parameters while holding Gaussian parameters fixed—crucial for precise scene alignment during the incremental stage. While the original implementation of 3DGS is not directly applicable to large-scale terrains due to computational and memory constraints, we have previously addressed these challenges by developing adaptive tiling strategies to partition and efficiently manage extensive 3D scenes, enabling successful large-scale terrain reconstruction (Chen et al., 2024). We will further detail our camera pose refinement methodology utilizing 3DGS in Camera pose estimation section.

**Data collection**
To support both the development and evaluation of the IDU pipeline, we collected distinct datasets from two different sites: one at the Simulation and Training Technology Center (STTC) in Orlando, Florida, and the other at the Geronimo CACTF in Fort Johnson, Louisiana. These datasets vary in complexity, terrain characteristic, and data acquisition methods, offering a comprehensive foundation for testing both individual modules and the complete IDU pipeline.

The STTC dataset was primarily used for developing and testing the core components of the IDU pipeline. On the first day of the collection, over 3,000 aerial images were captured around the STTC building using a UAV, providing sufficient coverage for reconstructing a high-resolution baseline 3D terrain model with Gaussian Splatting. On the second day, a limited number of ground-level images were collected after manually introducing changes, including pop-up tents and Ground Control Point (GPC) markers placed directly on the ground. This dataset offered a well-controlled environment for early-stage development, enabling us to evaluate and refine individual pipeline modules—such as camera pose estimation, change detection, and 3D object placement—under simplified and consistent conditions.

The Geronimo CACTF dataset was designed to test the full IDU pipeline in a more operationally diverse and large-scale environment. Data was collected using three different drone platforms with varying imaging capabilities. The baseline aerial imagery was collected using both a fixed-wing drone (eBee TAC) and a lightweight quadcopter (Anafi Parrot). The eBee TAC, equipped for long-endurance flights and wide-area coverage, was used to map the large-scale site efficiently and at high resolution. The quadcopter, capable of agile flight and tilt gimbal imaging, complemented this by capturing lower-altitude oblique and cross hatch views, enabling more detailed reconstructions on building facades and small objects. After capturing the data for the baseline 3DGS model, a series of physical changes were introduced in the environment, including placing sandbags for fortified fighting position, simulated munition stockpiles, dud bombs, road barriers, checkpoint booths, and pop-up tents at various locations throughout the area. To simulate a real-world update scenario, a third drone (Skydio X2D), equipped with close-range imaging capability was used to capture the post-change imagery. Unlike the platforms used for the initial mapping, the Skydio was tasked specifically with documenting the newly introduced objects from near ground levels. The key objective of this setup was to evaluate whether the IDU pipeline could integrate images from a different drone and sensor configuration than the one used to create the baseline scene. This reflects realistic field conditions, where updates are often captured opportunistically using different platforms. Figure 1 shows the 3DGS model for each of these datasets.

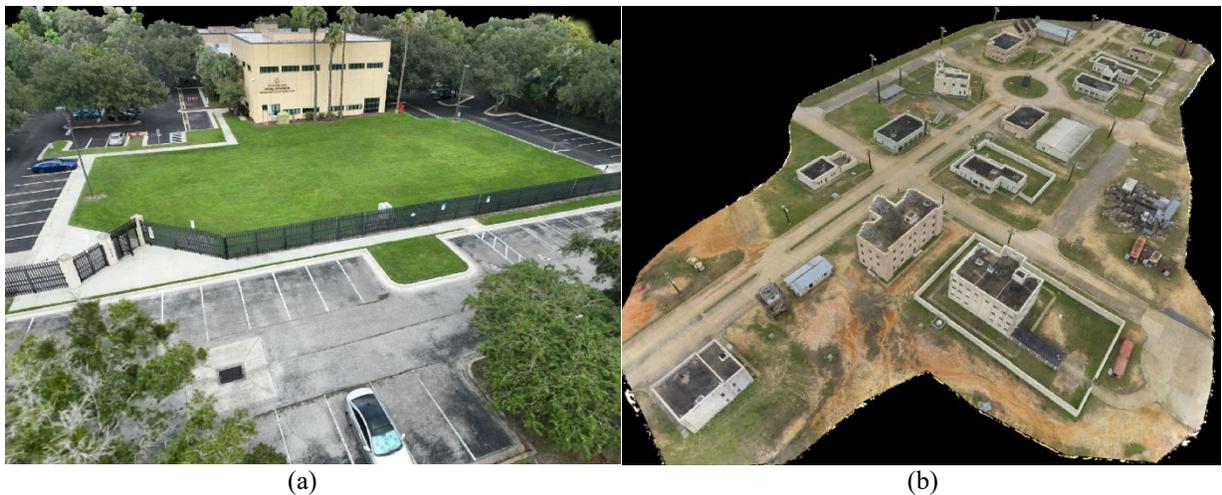

(a)  (b)
**Figure 1.** The baseline 3DGS models for the (a) STTC building and (b) Geronimo CACTF.

**Incremental Dynamic Update (IDU) pipeline**

**Overview of IDU**

Figure 2 presents an overview of our designed and implemented IDU pipeline, which addresses the challenge of maintaining high-fidelity 3D reconstructions in environments subject to frequent changes. The pipeline is organized into four main stages 3D Reconstruction, Incremental, Dynamic, and Update. It begins with the generation of a baseline 3D environment using multi-view aerial imagery processed with our previously developed large-scale 3D Gaussian Splatting (3DGS) pipeline. In this stage, a high-quality 3DGS model is created directly from the input images, capturing both geometric structure and photorealistic appearance. This model serves as the foundation for all subsequent computations and supports accurate rendering and alignment throughout the process. The modular design of the pipeline enables precise and efficient integration of new objects into the existing baseline without requiring full reconstruction.

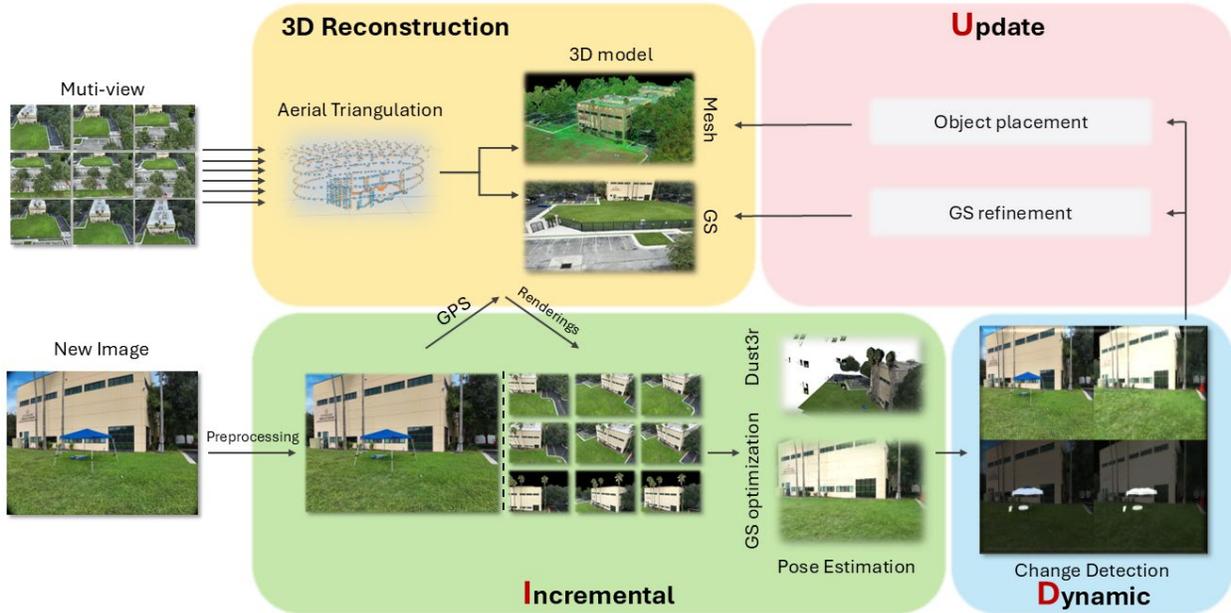

**Figure 2.** Overview of the IDU pipeline, the pipeline begins with an initial 3D Reconstruction, proceeds with an Incremental stage where new images are registered, then applies a Dynamic procedure to detect changes, and finally Updates and integrates the new 3D content.

In the **Incremental** stage, newly captured images reflecting recent changes are introduced into the system. These images are aligned with the baseline 3DGS model through a two-step camera pose estimation process. In the first step, we adapted DUSt3R (see Adaption of DUSt3R for Camera Pose Estimation section for details on DUSt3R) and its global alignment capability by incorporating rendered reference views from the existing 3D model, along with known camera poses and depth maps. Use of DUSt3R enables us to estimate the camera poses of new images relative to the current scene. Subsequently, these initial pose estimates are refined using a differentiable optimization procedure (discussed in 3D Gaussian Splatting based camera pose refinement section) that leverages the fixed parameters of the existing baseline 3DGS model. This refinement adjusts only the camera pose parameters, ensuring precise alignment of new images with the existing 3D environment.

Following the camera pose estimation process, in the **Dynamic** stage, we detect and isolate scene changes by comparing the newly aligned images against renderings generated from the existing baseline 3D model. For this task, we leveraged the C-3PO framework (C-3PO details discussed in Change detection section) to identify discrepancies between the new and rendered images. This process highlights objects that have been introduced since the baseline reconstruction, enabling targeted updates in the next stage.

Finally, in the **Update** stage, the identified scene changes are transformed into 3D assets using TRELLIS (details discussed in Generation and placement of new object section), a framework developed by Microsoft for generating high-quality 3D models from image inputs. These newly generated assets are then integrated into the existing 3DGS environment under human supervision, ensuring spatial accuracy and visual coherence. The following sections provide detailed technical descriptions for each step within this pipeline, including camera pose estimation change detection, and 3D model update strategies.

**Camera pose estimation (Incremental)**

**Synthetic image rendering for camera pose estimation**

To estimate the camera pose for a newly captured image, we begin by extracting its initial positional and orientation data from the image's metadata, including GPS coordinates and gimbal-based rotation angles provided by the UAV. This initial position and orientation usually are not accurate enough for later processes. Using this as an approximate reference, we render a set of synthetic views from the existing 3DGS model at nearby camera positions. These rendered synthetic views are generated using one of five complementary strategies—*simple*, *curve*, *hemisphere*, *quarter sphere,* and *overhead*—each designed to simulate different spatial relationships around the reference image's location. Each strategy introduces controlled variations in camera position and orientation to create different types of spatial and visual overlap with the reference image. The *simple* strategy perturbs the camera along axis-aligned directions while keeping the original view direction, making it suitable for open areas. The *curve* and *hemisphere* strategies place cameras along curved paths or domes centered around the reference image position, with each view both shifted and slightly rotated relative to the original pose, while maintaining the same general viewing direction. This provides broad angular coverage useful in complex scenes. The *quarter sphere* strategy restricts samples to the upper part of the hemisphere, helping avoid camera placements too close to the ground, particularly useful when the reference image is near ground level. The *overhead* strategy provides steep, downward-looking views suited for high-elevation images such as rooftop or open-ground observations. Figure 3 shows these five rendering strategies.

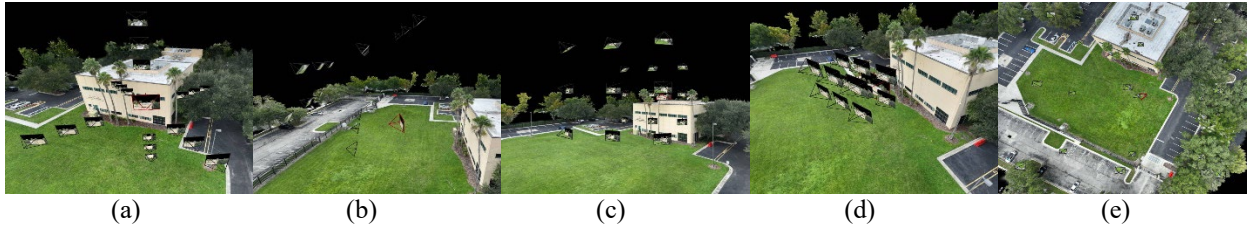

(a)            (b)            (c)            (d)            (e)

**Figure 3.** Rendering strategies: (a) simple, (b) curve, (c) hemisphere, (d) quarter sphere, and (e) overhead.

After rendering the synthetic images, we apply a filtering process to ensure only high-quality views are used for camera pose estimation. First, we remove any cameras that fall below the terrain surface by computing their height above-ground level (AGL) using a KD-tree constructed from the 3DGS point cloud. Views with height AGL less than a fixed threshold (e.g., 1 meter) are excluded to prevent visual occlusion or invalid geometry. Additionally, we discard rendered images that are likely to be uninformative—such as those dominated by empty space, overly close geometry, or excessive visual artifacts. This is done by analyzing the rendered depth maps and removing images that contain too many missing or overly near-depth values. These heuristics help eliminate viewpoints that are likely to be blurry or structurally ambiguous, improving both the reliability and accuracy of the downstream pose estimation.

Since all synthetic views are rendered directly from the 3DGS model, their camera intrinsics and poses are precisely known. These views, along with their corresponding depth maps, are paired with the new/reference image and passed into the pose estimation pipeline. By supplying known synthetic image camera poses and depth maps, we constrain the estimated pose of the new/reference image within the same coordinate reference frame of the existing 3DGS model.

**Adaption of DUSt3R for Camera Pose Estimation**

We base our camera pose estimation process on DUSt3R (Dense and Unconstrained Stereo 3D Reconstruction), a recent 3D vision foundational model (network) that introduces a unified framework for predicting dense scene geometry and supporting a range of downstream tasks including structure-from-motion, pose estimation, and global alignment (Wang, Leroy, Cabon, Chidlovskii, & Revaud, 2024). At the core of DUSt3R is a dense matching framework built on a shared Vision Transformer (ViT) encoder, which processes image pairs in a Siamese manner to extract token features that are then passed through cross-attending transformer decoders to regress dense 3D pointmaps. These pointmaps contain per-pixel 3D coordinates defined in view-specific local frames and serve as a generalizable geometric representation that eliminates the need for sparse key point detection, calibration, or precomputed depth.

This dense pointmap formulation enables DUSt3R to be applied flexibly to a variety of downstream tasks. One of these is global alignment, in which DUSt3R estimates the scale, rotation, and translation that align local pointmaps from multiple image pairs into a consistent global coordinate reference frame. In its original design, DUSt3R jointly optimizes both the global 3D pointmap and the per-view similarity transforms (scale, rotation, and translation) that register each local pointmap to the global structure. This formulation assumes no prior knowledge of camera poses or depth, making it suitable for fully unconstrained multi-view scenarios. However, the joint optimization can be ambiguous in scale when no prior structure is available.

In our setting, we adapt DUSt3R's global alignment module to operate in a partially constrained environment, where synthetic images are rendered based on an existing 3DGS model. These rendered views come with known camera poses and accurate depth maps, which we use to directly construct a fixed global pointmap. Instead of optimizing the global structure, we hold this constructed geometry constant and optimize only the similarity transform (scale, rotation, and translation) for the new/reference image's predicted pointmap. Our modification of DUSt3R is to include rendered synthetic images along with the known camera poses and depth maps as the input data.

Through this modification, DUSt3R produces a camera pose for the new/reference image that is aligned with the coordinate reference frame of the existing 3DGS model. More importantly, DUSt3R's global alignment mechanism is robust to scene changes due to its use of confidence scores which quantify the geometric reliability of each predicted 3D point. During optimization, only high-confidence points meaningfully influence the alignment, while low-confidence regions—such as those corresponding to new or changed objects that are not present in the baseline model—are naturally down weighted. This mechanism is critical in the IDU context: even when the new image includes added elements, DUSt3R effectively focuses the alignment on stable, overlapping regions shared with the reference geometry. As a result, our modified DUSt3R pipeline offers both structural consistency and change tolerance, enabling robust pose estimation in dynamic environments where incremental updates are essential. Figure 4 presents the source images containing newly added objects alongside the rendered views generated using the estimated camera poses from our modified DUSt3R pipeline.

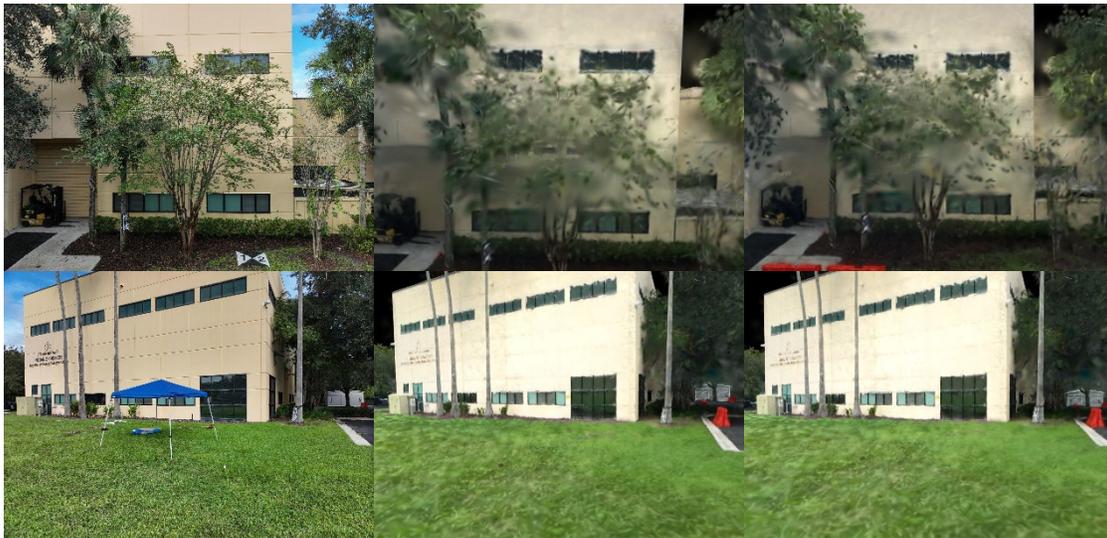

**Figure 4.** Camera pose estimation results for two reference images. Each row shows the source image (left column), rendered view using the DUSt3R-estimated pose (middle column), and rendered view using the 3DGS-refined pose (right column).

**3D Gaussian Splatting based camera pose refinement**

To achieve precise camera alignment beyond the initial estimation provided by our modified DUSt3R pipeline, we exploit the differentiable rendering capability inherent from the 3D Gaussian Splatting (3DGS) representation. We treat camera pose refinement explicitly as an optimization problem aimed at minimizing the rendering discrepancy between the actual captured image and the corresponding view rendered from the adjusted camera pose. Specifically, during this refinement stage, we hold all Gaussian splat parameters—including their spatial positions, scales, rotations, colors, and opacity—constant to preserve the integrity and computational efficiency of the baseline 3DGS model.

Our optimization module introduces incremental pose adjustments (deltas) to the initial camera poses estimated by DUSt3R. Specifically, these pose deltas are encoded using translation vectors and rotation parameters represented compactly in a 9-element tensor space. By applying these incremental adjustments, we iteratively refine the camera-to-world transformation matrices. At each optimization iteration, we render the 3DGS scene from the adjusted camera pose and compute a composite loss function based on the pixel-wise L1 differences and structural similarity index (SSIM) between the rendered image and the real image. This combined loss function balances pixel-level accuracy with perceptual quality, ensuring effective convergence to the correct pose. The differentiable rendering nature of 3DGS allows gradients derived from image differences to directly propagate back through the rendering process, enabling gradient-based optimization of the camera pose parameters. This iterative refinement results in precise convergence to the optimal camera alignment. Figure 4 shows examples of rendered views generated using the refined camera poses, demonstrating improved alignment from the DUSt3R results.

**Change detection (<u>Dynamic</u>)**

Once newly captured images have been accurately aligned with the existing 3D model, the next step within the IDU pipeline involves detecting and isolating scene changes. The simplest approach to change detection is manual: a human operator can directly place bounding boxes around the introduced object visible in the new imagery. Although this manual method is effective and reliable for scenarios with relatively few changes, it becomes impractical for large-scale or frequently changing environments.

Automated change detection methods, particularly semantic-based approaches, offer a promising solution for larger and more complex scenarios. Among these, the C-3PO (Combine 3 POssible change types) method represents a recent advancement by reframing change detection as a semantic segmentation problem (Wang, Gao, & Wang, 2023). Unlike traditional pixel-level change detection methods, which primarily rely on direct pixel differences and are thus vulnerable to errors caused by minor camera misalignments, C-3PO approaches change detection at a semantic level. It categorizes changes explicitly into three distinct semantic types—objects that "appear," "disappear," or "exchange"—and separately learns representations for each type.

The primary advantage of adopting a semantic segmentation paradigm like C-3PO lies in its immunity to minor inaccuracies in camera pose estimation. Pixel-level approaches can easily yield false positives or negatives when there is even slight misalignment between the newly captured image and the reference rendering. In contrast, semantic segmentation frameworks inherently tolerate small geometric discrepancies because they aggregate change signals at the object or semantic level rather than purely at the pixel level. As a result, they produce more stable and meaningful change detections in realistic operational scenarios, where perfect pixel alignment may be impractical. Figure 5 shows five examples of detected changes in the images collected at STTC.

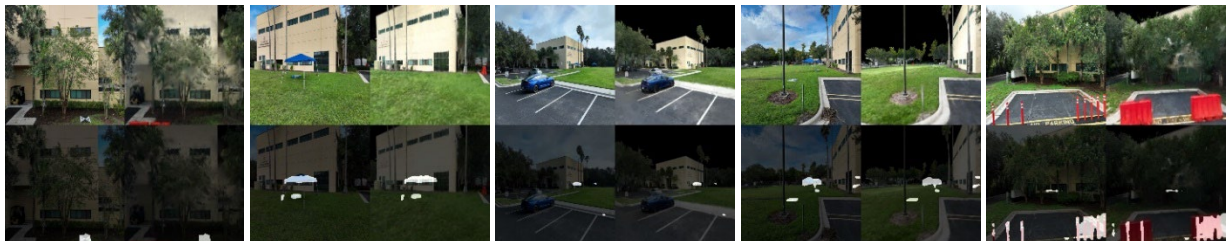

**Figure 5.** Illustration of change detection using C-3PO. For each sample, the top-left shows the image with changes, the top-right shows the rendered image using estimated camera poses, the bottom-left presents the detected changes over the source image (highlighted in white), and the bottom-right overlays the same change mask onto the rendered image.

**Generation and placement of new object (<u>Update</u>)**

Once the new object has been identified and isolated within the imagery from the Dynamic stage, the IDU pipeline moves to generating explicit 3D representation of the new object and integrating it into the existing 3D environment. In our current implementation, each detected change is included in a 2D single-view cropped image. Therefore, we specifically leverage TRELLIS (Xiang et al., 2024), a recently proposed generative framework designed to produce detailed and accurate 3D assets from limited image inputs, including single-view image.

TRELLIS generates high-quality 3D assets from single-view image through a two-stage generation process built on its structured latent representation. In the first stage, TRELLIS predicts a sparse 3D structure by generating a distribution of active voxel positions using a transformer trained with a rectified flow objective. To enable efficient training, a dense binary voxel grid is first compressed into a lower-resolution continuous feature grid, which is then denoised by a transformer conditioned on image features extracted from a pretrained vision model i.e., DINOv2 (Oquab et al., 2023). In the second stage, TRELLIS generates local latent codes for each active voxel location using a separate sparse transformer model. These latent codes encode local geometry and appearance details, allowing for high-resolution reconstruction. Similar to the structure generator, this latent code generator is also trained with rectified flow matching, and supports conditioning on image inputs. The final structured latent can be decoded into multiple 3D formats. In our IDU pipeline, we use the TRELLIS 3DGS decoder to directly generate assets compatible with our scene representation. TRELLIS's ability to produce accurate 3D outputs from a single image crop makes it particularly suited for incremental updates where only one view of a new object is available. Figure 6 shows the source images alongside the generated 3D assets created from their cropped views.

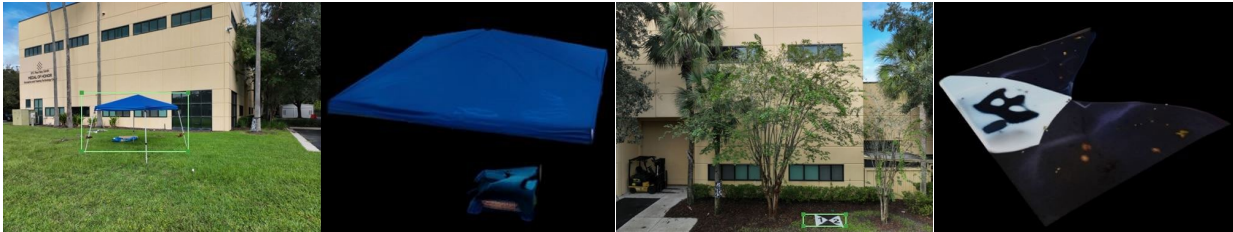

**Figure 6.** Single-view 3D asset generation using TRELLIS. The left pair shows a source image with detected changes (green bounding box) and the corresponding 3D model generated by TRELLIS. The right pair presents a second example following the same structure.

Once the 3D asset is generated by TRELLIS, it exists in its own canonical coordinate system and is not aligned to the real-world coordinate frame of the original image. This is because TRELLIS operates as a generative model and does not preserve the spatial relation between the generated object and the camera used to create the input crop. To integrate the new object into the existing 3D environment, we perform a post-generation alignment step that maps the object from its latent coordinate system into the global coordinate reference frame. To accomplish this, we implement a rough alignment procedure that leverages both the camera intrinsics and the depth map derived from the original scene via DUSt3R. Given the object mask in the input image, we compute the median depth of the object region and back-project the center of the mask into 3D camera coordinates. These coordinates are then transformed into world space using the known camera-to-world transformation. At the same time, we estimate the center of the generated splats and compute a translation that moves the asset into its intended scene location. In addition to translation, we apply uniform scaling to adjust the size of the asset to match the scale of real-world objects. This is achieved by comparing the bounding box extent of the generated splats to the bounding box of the object region extracted from the depth map.

**Research prototype implementation**

To evaluate the proposed IDU pipeline, we developed a modular research prototype with a graphical user interface that supports interactive scene updating, camera pose estimation, change detection, and 3D asset integration. The system is implemented in Python using PyTorch for core computations, PyQt for GUI functionality, and Viser for real-time visualization of 3D Gaussian Splatting environments.

A key feature of the prototype is the ability for users to interactively refine camera poses at multiple stages. The initial pose can be derived from GPS metadata, and then optionally refined through DUSt3R-based estimation or further improved using gradient-based optimization within the 3DGS rendering framework. Users may choose to initiate synthetic image rendering strategies—used for pose estimation—from any of these pose estimates, offering flexibility and robustness in diverse scenarios. Once camera pose alignment is finalized, the user selects the region of change within the image. This cropped image is passed to TRELLIS, which generates a 3D asset from a single view. After the asset is generated, the system applies the rough placement procedure to align the object into the global 3D environment. The prototype further supports manual adjustment of the object's position and scale within the environment, allowing users to correct for residual errors in alignment or geometry. These adjustments are applied in real time and visualized immediately within the 3DGS-rendered scene.

The prototype demonstrates the feasibility of our IDU pipeline and highlights the balance between automation and user control. It serves as a platform for experimentation with integration strategies and prepares the foundation for future work on automating the steps that require manual (but assisted) intervention, such as change detection and object pose refinement. A video demonstration of our IDU prototype can be found at: https://youtu.be/FXRrmco8gnU

**Case Studies (IDU for Geronimo CACTF)**

To evaluate the full pipeline under realistic operational conditions, we applied the IDU process to data collected from the Geronimo CACTF at Fort Johnson. As the intended end use of the 3D models in this setting is for training visualization and simulation, our evaluation emphasized visual quality and alignment fidelity in the updated scenes. The key objective was to determine whether small sets of newly acquired images could be used to incrementally and efficiently update the 3D virtual environment without requiring full reconstruction processing. For this test, only four ground-level images capturing new changes were selected and processed through the IDU pipeline. We visualized the results by rendering views of the updated 3D Gaussian Splatting scene and compared them directly with the source images used in the update. Figure 7 shows side-by-side renderings of the updated 3D scene and the real images used for IDU.

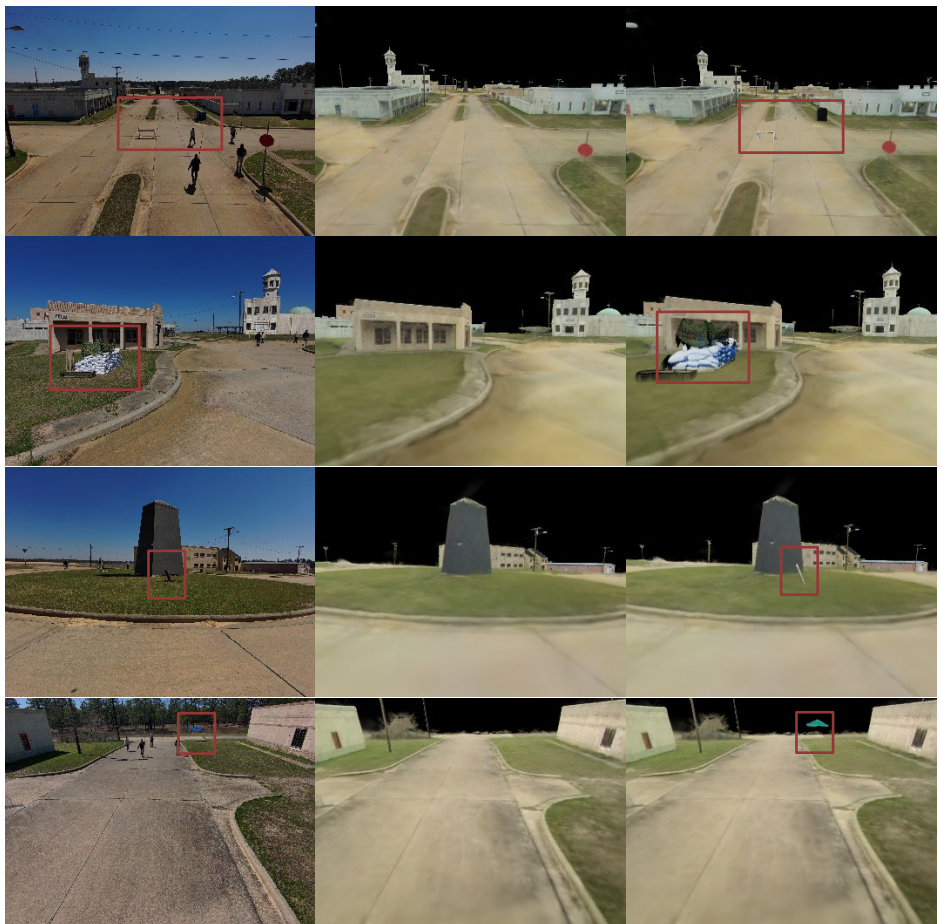

**Figure 7.** Visual comparison of the IDU results. Each row shows a source image with new changes (left), the corresponding rendered view from the baseline 3DGS model (middle), and the rendered view with the changes from IDU updated 3DGS model(right), changes are highlighted in red box.

During this evaluation, we observed that the 3DGS-based camera pose refinement did not perform reliably for certain images in this dataset. This limitation stemmed from the presence of large sky regions in the input images. Because the sky was not part of the baseline 3DGS reconstruction, the rendered reference views lacked matching features in those areas. As a result, the optimization often became stuck in local minima, particularly when the majority of the images contained unrepresented sky regions. In such cases, we were able to improve the camera pose by rerunning DUSt3R, this time using synthetic images rendered from the previously estimated pose to guide the alignment.

To further compare the efficiency of our approach against with traditional reconstruction methods, we measured the time required for both entire environment reprocessing and incremental updates. Reconstructing the entire environment from scratch, including full photogrammetric processing and manual ground control point (GCP) alignment, typically requires approximately 16 hours of labor and computation time. In contrast, updating the model using the IDU pipeline—including image acquisition and completed integration—takes less than one hour. This substantial reduction in turnaround time illustrates the practical benefits of adopting an incremental update approach for maintaining simulation-ready 3D environments. Table 1 summarizes the time cost comparison between full reconstruction and our incremental update strategy applied on Geronimo CACTF data.

|  | # of images | Data collection | Camera pose estimation | 3D model generation | Total |
| --- | --- | --- | --- | --- | --- |
| Full reconstruction | 6554 | 4 hours | 2 hours | 10 hours | 16 hours |
| IDU | 4 | 15 minutes | 20 minutes | 5 minutes | 40 minutes |

**Table 1.** Time comparison between full-scene reconstruction and the proposed IDU pipeline.

**Discussion, Limitations, and Conclusion**

This paper presented our designed Incremental Dynamic Update (IDU) pipeline, a practical solution for maintaining up-to-date 3D virtual environments in dynamic settings using minimal new imagery data. Through a modular design involving pipeline steps of camera pose estimation, change detection, and AI-driven 3D asset generation, IDU enables localized updates without reprocessing the entire environment. Our results demonstrate that IDU substantially reduces the time to keep simulation environments current while maintaining high visual fidelity. Experimental evaluations on two diverse datasets—a controlled environment at STTC and a more complex environment at Geronimo CACTF—validate the pipeline's effectiveness. In the STTC case, the modular components were iteratively developed and verified in isolation. With the Fort Johnson data, we tested the entire end-to-end system using imagery from multiple UAV platforms, highlighting IDU's flexibility to support cross-platform integration. The reduction in update time, from approximately 16 hours (with conventional reprocessing) to about one hour using our IDU pipeline, underscores IDU's value in real-world applications where timelines are essential.

While the IDU pipeline shows promising results, there are limitations and areas for further exploration. In particular, 3DGS-based camera refinement was shown to be sensitive to input images with large sky regions, due to the absence of sky data in the original reconstruction. Addressing this issue may involve extending the 3DGS baseline model to better represent the background content or incorporating prior information to regularize pose optimization under weak rendered image conditions. Additionally, although our current prototype supports user interaction for change selection and object placement, automating these steps through learned models remains an active area of research.

Our future work will focus on improving robustness in supporting multi-view asset fusion for more complete 3D generation, and integrating temporal reasoning for automated change detection across time-series imagery. Additionally, we plan to develop quantitative evaluation metrics, such as positional accuracy, reconstruction completeness, and rendering quality measures, to objectively assess the pipeline's performance. We also aim to evaluate the pipeline's scalability on larger environments and with higher-frequency updates, which are increasingly relevant for military use cases. In conclusion, the IDU pipeline offers a significant advancement in how virtual environments can be efficiently maintained and adapted. By combining modern neural rendering, camera pose estimation, and generative modeling, IDU approach provides a cost-effective and scalable path for sustaining immersive 3D simulations that remain aligned with the real world.

**ACKNOWLEDGEMENTS**
The authors would like to thank our primary sponsor of this research: Mr. Clayton Burford of the Battlespace Content Creation (BCC) team at Simulation and Training Technology Center (STTC). This work is supported by University Affiliated Research Center (UARC) award W911NF-14-D-0005. Statements and opinions expressed and content included do not necessarily reflect the position or the policy of the Government, and no official endorsement should be inferred.

# REFERENCES

Chen, M., Feng, A., McCullough, K., Prasad, P. B., McAlinden, R., Soibelman, L., & Enloe, M. (2019a). Fully automated photogrammetric data segmentation and object information extraction approach for creating simulation terrain. *Interservice/Industry Training, Simulation, and Education Conference (I/ITSEC)*.

Chen, M., Feng, A., McCullough, K., Prasad, P. B., McAlinden, R., & Soibelman, L. (2020a). Semantic segmentation and data fusion of microsoft bing 3d cities and small uav-based photogrammetric data. *Interservice/Industry Training, Simulation, and Education Conference (I/ITSEC)*.

Chen, M., Feng, A., McCullough, K., Prasad, P. B., McAlinden, R., & Soibelman, L. (2020b). Generating synthetic photogrammetric data for training deep learning based 3D point cloud segmentation models. *Interservice/Industry Training, Simulation, and Education Conference (I/ITSEC)*.

McCullough, K., Feng, A., Chen, M., & McAlinden, R. (2020). Utilizing satellite imagery datasets and machine learning data models to evaluate infrastructure change in undeveloped regions. *Interservice/Industry Training, Simulation, and Education Conference (I/ITSEC)*.

Chen, M., Feng, A., Hou, Y., McCullough, K., Prasad, P. B., & Soibelman, L. (2021a). Ground material classification for UAV-based photogrammetric 3D data A 2D-3D Hybrid Approach. *Interservice/Industry Training, Simulation, and Education Conference (I/ITSEC)*.

Xu, J., Chen, M., Feng, A., Yu, Z., & Shi, Y. (2024). Open-Vocabulary High-Resolution 3D (OVHR3D) Data Segmentation and Annotation Framework. *Interservice/Industry Training, Simulation, and Education Conference (I/ITSEC)*.

Chen, M., Feng, A., McAlinden, R., & Soibelman, L. (2020c). Photogrammetric point cloud segmentation and object information extraction for creating virtual environments and simulations. *Journal of Management in Engineering, 36(2), 04019046.*

Chen, M., McAlinden, R., Spicer, R., & Soibelman Ph D, L. (2019b). Semantic modeling of outdoor scenes for the creation of virtual environments and simulations.

Chen, M., Feng, A., McCullough, K., Prasad, P. B., McAlinden, R., & Soibelman, L. (2020d). 3D photogrammetry point cloud segmentation using a model ensembling framework. *Journal of Computing in Civil Engineering, 34(6), 04020048.*

Chen, M., Hu, Q., Yu, Z., Thomas, H., Feng, A., Hou, Y., ... & Soibelman, L. (2021b). Stpls3d: A large-scale synthetic and real aerial photogrammetry 3d point cloud dataset. *33rd British Machine Vision Conference (BMVC)*.

Spicer, R., McAlinden, R., Conover, D., & Adelphi, M. (2016). Producing usable simulation terrain data from UAS-collected imagery. *Interservice/Industry Training, Simulation, and Education Conference (I/ITSEC)*.

Chen, M., Lal, D., Yu, Z., Xu, J., Feng, A., You, S., ... & Shi, Y. (2024). Large-Scale 3D Terrain Reconstruction Using 3D Gaussian Splatting for Visualization and Simulation. *International Archives of the Photogrammetry, Remote Sensing and Spatial Information Sciences-ISPRS Archives*, *48*.

Liu, R., Sun, D., Chen, M., Wang, Y., & Feng, A. (2025). Deformable Beta Splatting. arXiv preprint arXiv:2501.18630.

Hu, Y., Liu, R., Chen, M., Beerel, P., & Feng, A. (2025). SplatMAP: Online Dense Monocular SLAM with 3D Gaussian Splatting. Proceedings of the ACM on Computer Graphics and Interactive Techniques, 8(1), 1-18.

Liu, R., Xu, R., Hu, Y., Chen, M., & Feng, A. (2024). Atomgs: Atomizing gaussian splatting for high-fidelity radiance field. 35th British Machine Vision Conference (BMVC).

Hou, Y., Chen, M., Feng, A., & Li, S. (2024). Gaussian Splatting-Based Thermographic Modeling for Building Envelope Energy Audits. In 2024 IEEE 3rd International Conference on Intelligent Reality (ICIR) (pp. 1-2). IEEE.

Kerbl, B., Kopanas, G., Leimkühler, T., & Drettakis, G. (2023). 3d gaussian splatting for real-time radiance field rendering. *ACM Trans. Graph.*, *42*(4), 139-1.

Barron, J. T., Mildenhall, B., Tancik, M., Hedman, P., Martin-Brualla, R., & Srinivasan, P. P. (2021). Mip-nerf: A multiscale representation for anti-aliasing neural radiance fields. In Proceedings of the IEEE/CVF international conference on computer vision (pp. 5855-5864).

Barron, J. T., Mildenhall, B., Verbin, D., Srinivasan, P. P., & Hedman, P. (2023). Zip-nerf: Anti-aliased grid-based neural radiance fields. In Proceedings of the IEEE/CVF International Conference on Computer Vision (pp. 19697-19705).

Martin-Brualla, R., Radwan, N., Sajjadi, M. S., Barron, J. T., Dosovitskiy, A., & Duckworth, D. (2021). Nerf in the wild: Neural radiance fields for unconstrained photo collections. In Proceedings of the IEEE/CVF conference on computer vision and pattern recognition (pp. 7210-7219).

Mildenhall, B., Srinivasan, P. P., Tancik, M., Barron, J. T., Ramamoorthi, R., & Ng, R. (2021). Nerf: Representing scenes as neural radiance fields for view synthesis. Communications of the ACM, 65(1), 99-106.

Garbin, S. J., Kowalski, M., Johnson, M., Shotton, J., & Valentin, J. (2021). Fastnerf: High-fidelity neural rendering at 200fps. In Proceedings of the IEEE/CVF international conference on computer vision (pp. 14346-14355).

Wang, S., Leroy, V., Cabon, Y., Chidlovskii, B., & Revaud, J. (2024). Dust3r: Geometric 3d vision made easy. In *Proceedings of the IEEE/CVF Conference on Computer Vision and Pattern Recognition* (pp. 20697-20709).

Wang, G. H., Gao, B. B., & Wang, C. (2023). How to reduce change detection to semantic segmentation. Pattern Recognition, 138, 109384.

Xiang, J., Lv, Z., Xu, S., Deng, Y., Wang, R., Zhang, B., ... & Yang, J. (2024). Structured 3d latents for scalable and versatile 3d generation. *arXiv preprint arXiv:2412.01506*.

Oquab, M., Darcet, T., Moutakanni, T., Vo, H., Szafraniec, M., Khalidov, V., ... & Bojanowski, P. (2023). Dinov2: Learning robust visual features without supervision. arXiv preprint arXiv:2304.07193.